\documentclass{article}

\PassOptionsToPackage{numbers, compress}{natbib}
 \usepackage[preprint]{neurips_2026}

\usepackage[utf8]{inputenc} % allow utf-8 input
\usepackage[T1]{fontenc}    % use 8-bit T1 fonts
\usepackage{hyperref}       % hyperlinks
\usepackage{url}            % simple URL typesetting
\usepackage{booktabs}       % professional-quality tables
\usepackage{amsfonts}       % blackboard math symbols
\usepackage{nicefrac}       % compact symbols for 1/2, etc.
\usepackage{microtype}      % microtypography
\usepackage{xcolor}         % colors
\usepackage{amsmath}
\usepackage{graphicx} 

\title{SuperFace: Preference-Aligned Facial Expression Estimation Beyond Pseudo Supervision}

\author{%
  Zejian Kang\thanks{Equal contribution.} \\
  Zhejiang University \\
  Westlake University \\
  \texttt{} \\
  \And
  Xuanyang Xu\textsuperscript{*} \\
  The Chinese University of Hong Kong, Shenzhen \\
  \texttt{} \\
  \And
  Wentao Yang\textsuperscript{*} \\
  Zhejiang University \\
  Westlake University \\
  \texttt{} \\
  \AND
  Kai Zheng \\
  Westlake University \\
  \texttt{} \\
  \And
  Yuanchen Fei \\
  Hunan University \\
  \texttt{} \\
  \And
  Hongyuan Zou \\
  Westlake University \\
  \texttt{} \\
  \AND
  Hui Shan \\
  Zhejiang University \\
  Shanghai Innovation Institute \\
  Westlake University \\
  \texttt{} \\
  \And
  Shuo Yang \\
  The University of Texas at Austin \\
  \texttt{} \\
  \And
  Xiangru Huang\thanks{Corresponding author  \texttt{huangxiangru@westlake.edu.cn}.} \\
  Westlake University \\
   \\
}

\begin{document}

\maketitle

\begin{figure*}[h]
    \centering
    \includegraphics[width=\textwidth]{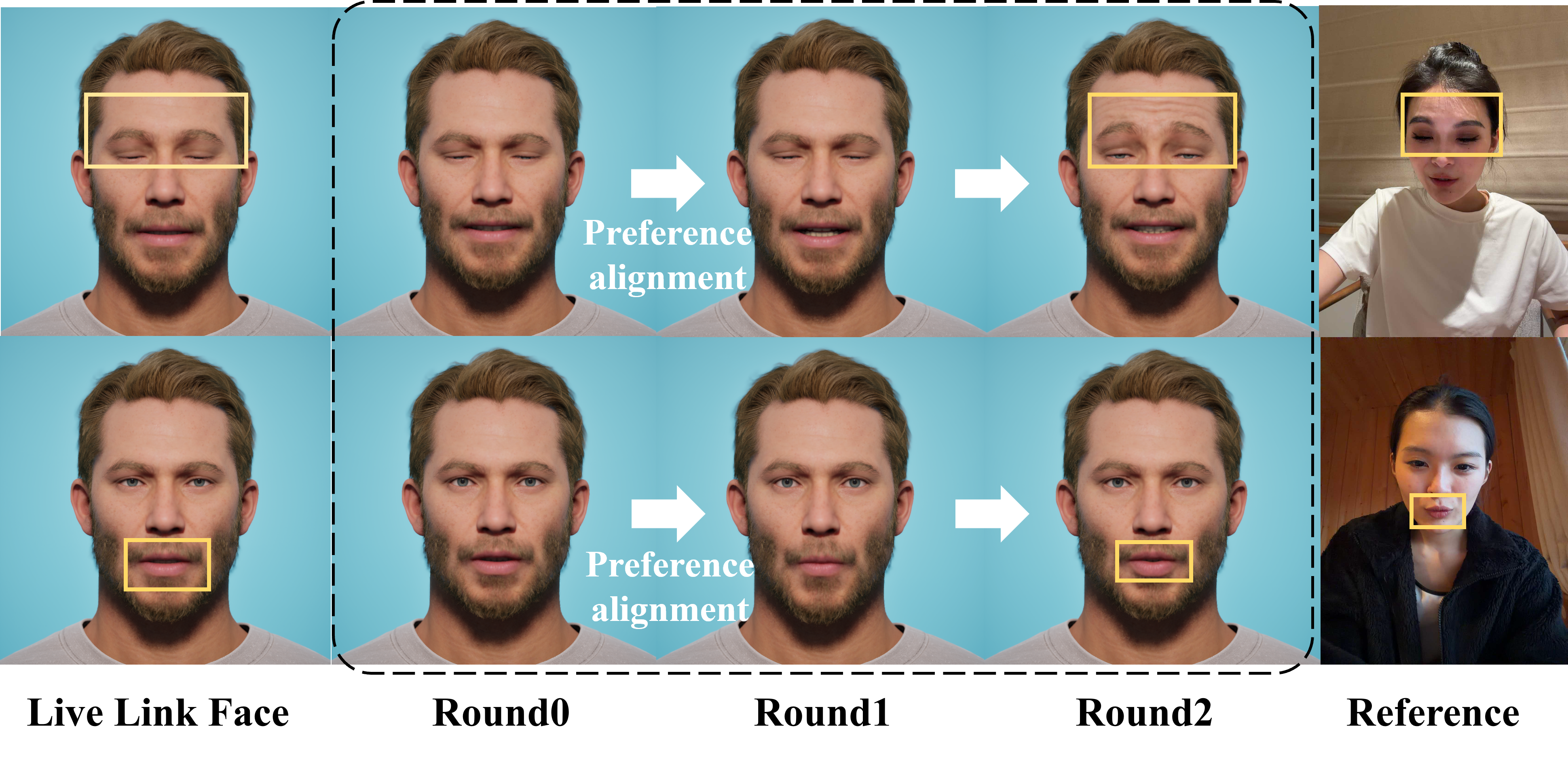}
    \caption{
    \textbf{SuperFace.}
    SuperFace improves ARKit facial expression estimation beyond pseudo supervision by incorporating human preference feedback on rendered results.
    Instead of treating Live Link Face coefficients as fixed ground truth, our method uses them as initialization, constructs region-aware preference comparisons for upper-face and lower-face motions, and iteratively optimizes the predictor toward perceptually preferred expressions.
    The examples show that SuperFace better captures localized facial actions such as eye and mouth movements, producing expressions that are more faithful to the reference images.
    }
    \label{fig:teaser}
\end{figure*}
\newpage
\begin{abstract}
Accurate facial estimation is crucial for realistic digital human animation, and ARKit blendshape coefficients offer an interpretable representation by mapping facial motions to semantic animation controls. However, learning high-quality ARKit coefficient prediction remains limited by the absence of reliable ground-truth supervision. Existing methods typically rely on capture software such as Live Link Face to provide pseudo labels, which may contain noisy activations, biased coefficient magnitudes, and missing or inaccurate facial actions. Consequently, models trained with supervised learning tend to reproduce imperfect pseudo labels rather than optimize for perceptual expression fidelity. In this paper, we propose \textbf{SuperFace}, a preference-driven framework that moves ARKit facial expression estimation from pseudo-label imitation toward human-aligned perceptual optimization. Instead of treating software-estimated coefficients as fixed ground truth, SuperFace uses them only as an initialization and further improves coefficient prediction through human preference feedback on rendered facial expressions. By aligning the model with perceptual judgments rather than numerical pseudo labels, SuperFace enables more visually faithful and expressive facial animation. Experiments show that SuperFace improves expression fidelity over Live Link Face supervision, demonstrating the effectiveness of preference-driven optimization for semantic facial action prediction.
\end{abstract}

\section{Introduction}
\label{sec:intro}

Facial expression capture is a fundamental task for digital human animation, with broad applications in virtual production, gaming, telepresence, and AR/VR. 
The perceived realism and expressiveness of a digital character are highly sensitive to subtle facial motions, making accurate expression capture essential for producing visually faithful facial animation~\cite{lewis2014practice,metahuman}.

Facial expression representations have evolved from geometry-centric parameterizations to semantically structured animation controls. 
Early methods relied on parametric face models and 3D morphable models (3DMMs)~\cite{park1974parametric,blanz1999morphable,egger2020morphable}, while later models such as FLAME improved facial modeling by disentangling identity, expression, and pose~\cite{li2017flame,feng2021deca,danvevcek2022emoca}. 
Although effective for geometric reconstruction, these representations are mainly compact deformation spaces and lack explicit correspondence to interpretable facial actions.

ARKit blendshape coefficients provide a more semantic and animation-friendly alternative~\cite{apple_arkit,lewis2014practice,grishchenko2023blendshapes}. 
Each coefficient corresponds to a predefined facial action, such as blinking, brow raising, or mouth opening, making the representation interpretable and compatible with real-time animation pipelines. 
This semantic structure also makes ARKit coefficients naturally suitable for multimodal large language models, which can associate visual facial cues with meaningful expression controls through vision-language priors~\cite{radford2021learning,li2023blip,liu2023llava}.

Despite the practicality of ARKit coefficients, accurate coefficient prediction remains challenging due to the lack of reliable ground-truth supervision. 
In practice, existing methods use coefficients estimated by capture software (Live Link Face) as training targets~\cite{apple_livelinkface_2025,menzel2022automated}. 
However, these coefficients are only pseudo labels and may contain noise, biased activation magnitudes, and missing or inaccurate facial actions, especially for subtle or ambiguous expressions. 
As a result, supervised fine-tuning on such targets mainly encourages the model to imitate the imperfect software outputs, rather than correct their errors. 
This limitation can propagate annotation noise into the learned model and lead to degraded expression fidelity, where the predicted coefficients may produce facial animations that are numerically close to the pseudo labels but visually inconsistent with the input expression.

To address these issues, we propose a preference-alignment framework for ARKit coefficient prediction. 
The model is first initialized through supervised fine-tuning with Live Link Face pseudo labels, and is then further refined through preference learning guided by human feedback over rendered facial expressions~\cite{rafailov2023direct}. 
We first introduce a region-aware preference construction strategy that decomposes ARKit coefficients into upper-face and lower-face subsets. 
Since a prediction may improve one facial region while degrading another, full-face annotation can be ambiguous. 
We therefore recombine predicted coefficients with pseudo ground-truth coefficients to isolate regional expression changes and obtain more fine-grained comparisons.

We further learn a human-aligned preference discriminator from user study annotations. 
Although large MLLMs provide strong visual-semantic priors~\cite{liu2023llava,qwen3vl}, they are not sufficiently reliable for judging subtle expression differences. 
The learned preference discriminator predicts human preferences between rendered expressions and guides preference optimization toward visually faithful ARKit predictions, rather than imitation of noisy Live Link Face outputs.

Our contributions can be summarized as follows:
\begin{itemize}
    \item We propose \textbf{SuperFace}, a preference-driven framework that improves ARKit facial expression estimation beyond fixed Live Link Face supervision, producing results better aligned with human perception.

    \item We introduce a \textbf{region-aware preference construction} strategy that decomposes full-face comparison into upper- and lower-face judgments, enabling more reliable pairwise preference annotation.

    \item We train an \textbf{MLLM-based preference discriminator} from initial-round human labels, enabling scalable preference supervision in later iterations.

    \item We optimize the policy with \textbf{DPO-based preference learning}, allowing the model to iteratively refine ARKit coefficients toward perceptually preferred facial actions.
\end{itemize}

\section{Related Work}

We organize related work into four aspects: face modeling, facial expression estimation, multimodal large models for facial expression understanding, and reinforcement learning for multimodal large models. We focus on how facial representations evolve from geometry-centric models to semantically structured animation controls, and how preference optimization can further improve perception-oriented facial prediction.

\paragraph{Face Modeling}

Facial modeling has long benefited from parametric representations, which reduce the complexity of high-resolution mesh modeling while preserving expressive capacity. Early 3D Morphable Models (3DMMs)~\cite{blanz1999morphable} represent facial geometry using linear PCA bases, and later models such as FLAME~\cite{li2017flame} provide compact and effective parameterizations for facial shape, expression, and pose. However, these representations are primarily geometry-centric deformation spaces and do not explicitly correspond to interpretable facial actions. In contrast, animation-oriented representations such as ARKit~\cite{apple_arkit} and MHR~\cite{ferguson2025mhr} define more semantically meaningful facial controls. In particular, ARKit blendshape coefficients are directly associated with predefined facial actions and can be conveniently captured using smartphone-based systems~\cite{menzel2022automated,Aloni_2025_ICCV,wu2025keyframefacetextexpressivefacial}, making them well suited for structured facial expression estimation and animation-ready prediction.

\paragraph{Facial Expression Estimation}

Early studies on 3D facial expression analysis relied mainly on explicit geometric cues such as landmarks, depth, mesh deformation, or morphable face models~\cite{ramanathan2006human,moeini20162d,lv20193d}. With the development of deep learning, CNN-based approaches began to directly regress expression-related 3D representations from images. For example, ExpNet~\cite{chang2018expnet} predicts 3D expression coefficients from image intensities, while 3DMCNN~\cite{jin2018learning} performs expression recognition directly on 3D facial meshes. More recent methods further incorporate reconstructed 3D faces or structured architectures for stronger facial prediction, including Ig3D~\cite{dong2024ig3d} and AUFART~\cite{kim2024action}. 

Another related direction focuses on reconstructing or estimating facial representations from monocular images, including ARKit-based and FLAME-based methods such as DeadFace~\cite{deadface2023}, SMIRK~\cite{retsinas2024smirk}, EMOCA~\cite{danvevcek2022emoca}, pixel3DMM~\cite{giebenhain2025pixel3dmm}, and SemanticFace~\cite{kang2026semanticface}. These methods demonstrate the importance of robust facial representation estimation, but many still rely on geometric fitting, face detection, or regression-style prediction. In comparison, semantically structured ARKit coefficients provide a more interpretable output space for facial action estimation, especially when combined with high-level semantic reasoning.

\paragraph{Multimodal Large Models for Facial Expression Understanding}

Recent studies have explored multimodal large models (MLLMs) for facial expression understanding. One line of work develops face-specific MLLMs or face-centric multimodal frameworks. FaceLLM~\cite{shahreza2025facellm} adapts MLLMs to face-related tasks through weakly supervised face-text data, FEALLM~\cite{hu2025feallm} introduces action-unit-aware reasoning for facial emotion analysis, and FacePhi~\cite{zhao2024facephi} uses facial landmarks as lightweight multimodal input for efficient facial understanding. Another line introduces language supervision or text-guided alignment into facial expression recognition and 3D/4D facial analysis~\cite{ma2025multimodal,behzad2025facet,li2025emoverse}. These works show that semantic priors and language-based reasoning can improve facial understanding. However, most existing methods focus on 2D recognition or high-level affective reasoning, while the integration of MLLM priors with structured 3D facial action coefficients remains underexplored.

\paragraph{Reinforcement Learning for Multimodal Large Models}

Reinforcement learning and preference optimization have become important post-training strategies for improving large models. PPO~\cite{schulman2017proximal} is a classical RLHF method, but it requires reward modeling, value estimation, and on-policy sampling, making it costly for large multimodal systems. DPO~\cite{rafailov2023direct} simplifies preference optimization by directly learning from chosen--rejected pairs without explicit reward modeling or online reinforcement learning, which makes it attractive for perception-oriented tasks where pairwise preferences are easier to collect. More recently, GRPO~\cite{shao2024deepseekmath} replaces critic-based value estimation with group-relative advantage estimation and has been applied to reasoning-oriented multimodal post-training, such as Vision-R1~\cite{huang2025vision} and R1-VL~\cite{zhang2025r1}. Compared with PPO and GRPO, DPO provides a simpler and more stable framework for facial expression prediction, where preference feedback can directly reflect perceptual quality of generated facial actions.

\begin{figure*}[t]
    \centering
    \makebox[\textwidth][c]{%
        \includegraphics[width=1.0\textwidth]{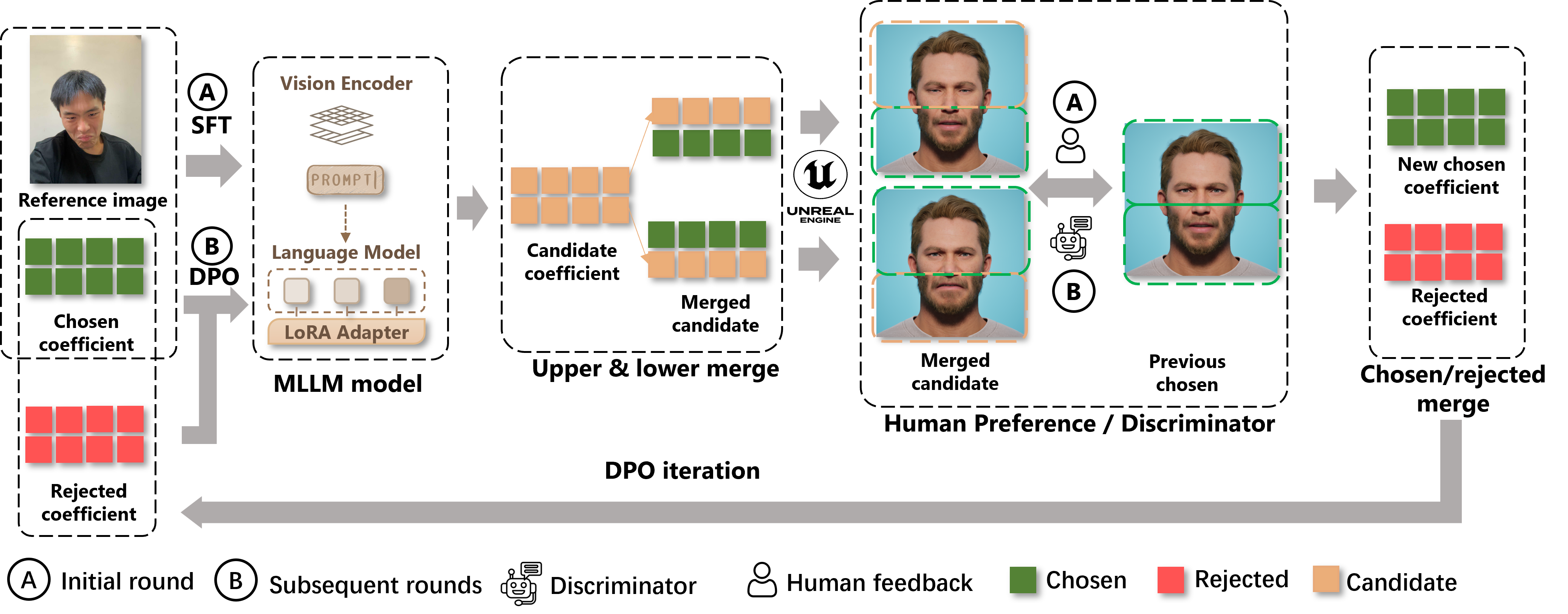}
    }
    \caption{\textbf{Overview of the proposed SuperFace pipeline.}
    SuperFace starts from supervised initialization on chosen ARKit coefficients, which are initialized by Live Link Face in the first round. The policy model then generates candidate coefficients, which are combined with the current chosen coefficients through upper-face and lower-face mixing to construct localized comparison pairs. Human annotators provide preference labels in the initial round, while a learned preference discriminator is used in subsequent rounds. The resulting chosen and rejected coefficient pairs are used for DPO-based iterative optimization.}
    \label{fig:method}
\end{figure*}

\section{Method}
Given a facial image $I \in \mathbb{R}^{H \times W \times 3}$, our goal is to predict a structured ARKit coefficient representation, which consists of predefined facial actions and their corresponding coefficient values. Specifically, we denote the facial action set as $\mathcal{A} = \{a_k\}_{k=1}^{K}$, where each $a_k$ represents a semantic facial action and $K=61$ in our setting. For each action $a_k$, the MLLM predicts a scalar coefficient $v_k \in \mathbb{R}$ that indicates its activation level. The facial expression estimation task is thus represented as
\begin{equation}
I \rightarrow{} \mathcal{S} = \{(a_k, v_k)\}_{k=1}^{K}.
\end{equation}

We propose \textbf{SuperFace}, a preference-driven framework for facial expression estimation in the interpretable ARKit coefficient space. Instead of treating Live Link Face coefficients as fixed ground truth, SuperFace uses them as an initial reference and iteratively improves the model through perceptual preference optimization. First, we adapt an MLLM to the ARKit action space by supervised fine-tuning (SFT) it with pseudo labels obtained from Live Link Face, enabling image-conditioned generation of serialized ARKit coefficients. Second, Sec.~\ref{subsec:preference_construction} constructs region-aware preference pairs by comparing policy-generated candidates with the current chosen coefficients through upper- and lower-face mixing, where initial-round human labels are further used to train a preference discriminator for later iterations. Finally, Sec.~\ref{subsec:dpo_training} optimizes the policy model with DPO using the constructed chosen-rejected coefficient pairs, progressively shifting the model toward perceptually preferred facial actions. An overview of our method is illustrated in Fig.~\ref{fig:method}.

% \subsection{Initialization with Supervised Fine Tuning}
% \label{subsec:sft_init}
% In this section, we learn an initial correspondence between facial images and ARKit coefficients through supervised fine tuning. Although the MLLM provides strong visual semantic priors for facial understanding, it is not explicitly aligned with the ARKit action space. To bridge this gap, we use ARKit coefficients captured by Live Link Face (LLF) as pseudo supervision, enabling the model to learn the basic image to coefficient correspondence and providing a reliable initialization for subsequent reinforcement learning. 

% Specifically, the loss function for the initial MLLM training is defined as
% \begin{equation}
% \mathcal{L}_{\text{SFT}} = \frac{1}{N}\sum_{i=1}^{N} \ell_{sft}\left(M_{\theta_{sft}}(I_i), \mathcal{S}^{LLF}_i\right),
% \end{equation}
% where $M_{\theta_{sft}}$ denotes the MLLM predictor parameterized by $\theta_{sft}$, and $\ell_{sft}$ is the token level negative log likelihood loss for autoregressive language modeling, while $\mathcal{S}^{LLF}$ $\mathcal{S}^{LLF}$ denotes the structured ARKit coefficient set captured by LLF.

\subsection{Region-Aware Human Preference Modeling}
\label{subsec:preference_construction}
In this section, we introduce the human preference modeling module, which serves as the reward function for subsequent reinforcement learning. Considering the distinct characteristics of different facial regions, we therefore organize this section into three parts: region-aware preference formulation, human preference annotation, and human preference discriminator training.

\paragraph{Region-Aware Preference Formulation.}
Given a reference image $I$ and two candidate coefficient sets $\mathcal{S}_{A}$ and $\mathcal{S}_{B}$, the module outputs a perceptual preference triplet that separately identifies the better match for the upper- and lower-face regions of $I$. Firstly, we decompose the structured action coefficient set as
\begin{equation}
\mathcal{S} = \mathcal{S}_u \cup \mathcal{S}_l ,
\end{equation}
where $\mathcal{S}_u$ and $\mathcal{S}_l$ denote the upper-face and lower-face coefficient subsets, respectively. Then, we construct two hybrid coefficient sets,
\begin{equation}
\mathcal{S}_{AB}=\mathcal{S}_{A,u} \cup \mathcal{S}_{B,l}, \qquad
\mathcal{S}_{BA}=\mathcal{S}_{B,u}\cup \mathcal{S}_{A,l}.
\end{equation}
Therefore, for each reference image $I$, we construct two region-wise comparison triplets:
\begin{equation}
\mathcal{T}_{u}=\{I, \mathcal{S}_{A},\mathcal{S}_{BA}\},
\qquad
\mathcal{T}_{l}=\{I, \mathcal{S}_{A},\mathcal{S}_{AB}\}.
\end{equation}
Here, $\mathcal{T}_{u}$ varies only the upper-face subset, while $\mathcal{T}_{l}$ varies only the lower-face subset, allowing region-wise preferences to be isolated. We define a human preference discriminator $\mathcal{D}$ to select the chosen and rejected regional subsets from each region-controlled comparison pair:
\begin{equation}
\left( I, \mathcal{S}_r^+, \mathcal{S}_r^- \right) = \mathcal{D} \left( \mathcal{T}_{r}  \right), \qquad r\in\{u,l\},
\label{eq:discirminator}
\end{equation}
where $\mathcal{S}_{r}^{+}$ denotes the regional coefficient subset that better matches the corresponding facial region of $I$, while $\mathcal{S}_{r}^{-}$ denotes the less preferred alternative. The region-wise outputs are then merged into the final human preference triplet:
\begin{equation}
\mathcal{T}_\mathcal{D}
=
(I,\mathcal{S}^{+},\mathcal{S}^{-}),
\quad
\text{where }
\mathcal{S}^{\pm}
=
\{\mathcal{S}_{u}^{\pm},\mathcal{S}_{l}^{\pm}\}.
\end{equation}

\paragraph{Human Preference Annotation.}

To align the reward model with human preferences, we collect a human preference dataset. Given a reference image $I$, we compare the facial image $I_M$, rendered from the ARKit coefficient prediction $\mathcal{S}_M$ produced by the SFT-adapted MLLM, with the facial image $I_{LLF}$, rendered from the Live Link Face coefficients $\mathcal{S}_{LLF}$. Following the region-aware preference formulation, we construct two region-wise comparison triplets:
\begin{equation}
\mathcal{T}_{H,u}=(I,\mathcal{S}_{LLF},\mathcal{S}_{ML}),\qquad\mathcal{T}_{H,l}=(I,\mathcal{S}_{LLF},\mathcal{S}_{LM}),
\end{equation}
where $\mathcal{S}_{ML}$ and $\mathcal{S}_{LM}$ denote recombined ARKit coefficient sets formed by exchanging the upper- and lower-face subsets between $\mathcal{S}_{M}$ and $\mathcal{S}_{LLF}$. Specifically, $\mathcal{T}_{u}^{H}$ isolates the upper-face difference, while $\mathcal{T}_{l}^{H}$ isolates the lower-face difference.
Given these two triplets, annotators select the candidate that better matches the corresponding facial region of the reference image, producing the chosen and rejected regional subsets:
\begin{equation}
(I,\mathcal{S}_{r}^{+},\mathcal{S}_{r}^{-})
=
\mathcal{D}^{H}\left(\mathcal{T}_{H,r}\right),
\qquad r\in\{u,l\}.
\end{equation}
The region-wise annotations are then merged into the human-annotated preference triplet:
\begin{equation}
\mathcal{T}_{H}
=
(I,\mathcal{S}^{+},\mathcal{S}^{-}),
\qquad
\mathcal{S}^{\pm}
=
\{\mathcal{S}_{u}^{\pm},\mathcal{S}_{l}^{\pm}\}.
\end{equation}
 For each reference image $I$, we collect pairwise preference annotations from multiple annotators. To alleviate annotation noise, we optionally incorporate bidirectional or repeated comparisons. Specifically, the same candidate pair can be presented in reversed order or assessed by multiple annotators, and only comparisons with consistent judgments are preserved. This results in cleaner win-lose pairs for preference optimization.

\paragraph{Preference Discriminator.}
After collecting the human preference dataset, we train a preference discriminator to predict human-aligned preferences between two candidate ARKit coefficient sets. 
We adopt the same MLLM architecture as in the SFT stage and repurpose it as a discriminator. 
For each reference image, the discriminator is applied to the two region-controlled comparison triplets, $\mathcal{T}_{u}$ and $\mathcal{T}_{l}$, and predicts the corresponding preference triplets following Eq.~\ref{eq:discirminator}. 
The human-annotated triplets $\mathcal{T}_{u}^{H}$ and $\mathcal{T}_{l}^{H}$ provide the target sequences. The discriminator is optimized using the same token-level negative log-likelihood objective as in the SFT stage:
\begin{equation}
\mathcal{L}_{\mathrm{dis}}
=
\frac{1}{N}
\sum_{i=1}^{N}
\sum_{r\in\{u,l\}}
\ell_{\mathrm{NLL}}\!\left(
\mathcal{D}\left( \mathcal{T}_{r}^{i} \right),
\mathcal{T}_{H, r}^{i}
\right),
\end{equation}
where $\mathcal{D}\left( \mathcal{T}_{r}^{i} \right)$ denotes the discriminator prediction for the $i$-th comparison triplet in region $r$, and $\mathcal{T}_{H, r}^{i}$ is the corresponding human-annotated preference triplet. After training, the discriminator predictions for the upper- and lower-face comparisons are merged into a coefficient-level preference triplet:
\begin{equation}
\mathcal{T}_\mathcal{D}
=
(I,\mathcal{S}^{+},\mathcal{S}^{-}),
\qquad
\mathcal{S}^{\pm}
=
\{\mathcal{S}_{u}^{\pm},\mathcal{S}_{l}^{\pm}\}.
\end{equation}
The predicted triplet is used for reward function that guides human preference optimization.

% After collecting the human preference dataset, we train a preference discriminator to predict human-aligned preferences between two candidate ARKit coefficient sets. 
% We adopt the same MLLM architecture as in the SFT stage and repurpose it as a discriminator. 
% For each annotated comparison sample, the discriminator predicts a preference triplet following Eq.~\ref{eq:discirminator}, where the human-annotated triplet $\mathcal{T}_{H}$ provides the target sequence.

% The discriminator is optimized using the same token-level negative log-likelihood objective as in the SFT stage:
% \begin{equation}
% \mathcal{L}_{\mathrm{dis}}
% =
% \frac{1}{N}
% \sum_{i=1}^{N}
% \ell_{\mathrm{NLL}}\!\left(
% \mathcal{D}(\mathcal{T}^{(i)}),\mathcal{T}_{H}^{(i)}
% \right).
% \end{equation}

% Consistent with the region aware annotation process, the discriminator is applied twice for each input image using two candidate pairs. This two step discrimination strategy provides richer region level preference supervision and improves the accuracy of the discriminator. Accordingly, each reference image yields two coefficient level preference triplets,
% \begin{equation}
% \mathcal{T}_{S}^{p} = (I, \mathcal{S}^{+}, \mathcal{S}^{-}),
% \end{equation}
% which are consistent with the preference triplets defined in the annotation stage and are used for subsequent human preference optimization.

\subsection{Preference Optimization}
\label{subsec:dpo_training}

After supervised fine-tuning (SFT), we obtain an MLLM-based facial action predictor $M_{\theta_{\mathrm{sft}}}$ with parameters $\theta_{\mathrm{sft}}$. 
We initialize the trainable policy model with $\theta^{(0)}=\theta_{\mathrm{sft}}$ and keep a frozen reference model with parameters $\bar{\theta}=\theta_{\mathrm{sft}}$. 
At the $k$-th preference optimization round, the policy model is denoted as $M_{\theta^{(k)}}$, while the reference model $M_{\bar{\theta}}$ remains fixed. 
For simplicity, we write $\pi_k(\mathcal{S}\mid I)$ for the policy distribution induced by $M_{\theta^{(k)}}$, and $\pi_{\mathrm{ref}}(\mathcal{S}\mid I)$ for the frozen reference distribution.

\paragraph{Preference Optimization Objective.}
At the $k$-th optimization round, the current policy generates candidate coefficient sets, from which the human preference discriminator constructs a coefficient-level preference triplet
\begin{equation}
\mathcal{T}_\mathcal{D}^{(k)}
=
(I,\mathcal{S}^{+},\mathcal{S}^{-}),
\qquad
\mathcal{S}^{\pm}
=
\{\mathcal{S}_{u}^{\pm},\mathcal{S}_{l}^{\pm}\}.
\end{equation}
Given this triplet, we optimize the policy model using Direct Preference Optimization:
\begin{equation}
\mathcal{L}_{\mathrm{DPO}}^{(k)}
=
-\log \sigma
\left(
\beta
\left[
\log \frac{\pi_{k}(\mathcal{S}^{+}\mid I)}
{\pi_{\mathrm{ref}}(\mathcal{S}^{+}\mid I)}
-
\log \frac{\pi_{k}(\mathcal{S}^{-}\mid I)}
{\pi_{\mathrm{ref}}(\mathcal{S}^{-}\mid I)}
\right]
\right),
\end{equation}
where $\sigma(\cdot)$ denotes the sigmoid function and $\beta$ controls the strength of preference optimization. 
Only the policy model is updated, while the reference model remains fixed.

\paragraph{Iterative Optimization.}
The preference optimization is performed iteratively. 
Starting from $\theta^{(0)}=\theta_{\mathrm{sft}}$, each round consists of three steps: candidate coefficient sets are first sampled from the current policy $\pi_k$, preference triplets are then constructed by the human preference discriminator, and the policy model is finally updated using the DPO objective. 
After each round, the updated policy is used to resample candidates and rebuild preference triplets for the next round. 
Through repeated sampling, preference discrimination, and DPO updating, the policy is progressively aligned with human perceptual preferences.

\paragraph{Stopping Criterion.}
We use a held-out evaluation set to monitor the optimization process. 
For each evaluation sample, we compare the rendering from the current policy prediction with the rendering from Live Link Face and compute the win rate:
\begin{equation}
\mathrm{WinRate}^{(k)}
=
\frac{1}{N}
\sum_{n=1}^{N}
\mathbb{I}
\left[
\mathcal{S}_{M}^{(k,n)}
\succ
\mathcal{S}_{LLF}^{(n)}
\right],
\end{equation}
where $N$ denotes the number of evaluation samples and $\mathbb{I}[\cdot]$ is the indicator function. 
Training terminates once the win rate exceeds a predefined threshold.

\section{Experiments}
\label{sec:experiments}

We conduct experiments to evaluate SuperFace from four aspects.
Sec.~\ref{subsec:exp_setup} describes the dataset, implementation details, baselines, and evaluation protocol.
Sec.~\ref{subsec:main_comparison} compares SuperFace with expression estimation methods under human preference evaluation.
Sec.~\ref{subsec:discriminator_eval} evaluates different preference discriminators and justifies using the trained MLLM-based discriminator to replace repeated human annotation.
Sec.~\ref{subsec:dpo_iter} analyzes the effect of iterative DPO optimization.

\subsection{Experimental Setup}
\label{subsec:exp_setup}

\paragraph{Dataset.}
We use the public ARKit facial expression dataset released by KeyframeFace~\cite{wu2025keyframefacetextexpressivefacial}, and follow the same preprocessing pipeline as SemanticFace~\cite{kang2026semanticface}.
The data are divided by subject into three subsets for different stages of the pipeline.
We use one subject with 3{,}022 frames for cold-start supervised fine-tuning (SFT), seven subjects with 31{,}231 frames for candidate rollout and preference data construction, and two held-out subjects with 9{,}664 frames for final evaluation.

\paragraph{Baselines.}
We compare our method with three baselines: \textbf{Live Link Face}~\cite{apple_livelinkface_2025}, \textbf{DeadFace}~\cite{deadface2023}, and \textbf{SemanticFace}~\cite{kang2026semanticface}.
Live Link Face is Apple's ARKit-based facial capture system, which leverages 3D structured light sensing to estimate facial geometry.
It serves as the initial pseudo ground truth in our pipeline.
Unlike pure 2D RGB-based approaches, the incorporation of depth information provides Live Link Face with higher accuracy and greater robustness in challenging lighting conditions.
DeadFace is an open-source MediaPipe-based method for facial geometry estimation from monocular images.
SemanticFace is a recent MLLM-based approach that predicts interpretable ARKit coefficients through semantic distillation.
Together, these baselines represent three different paradigms: industrial facial capture with structured-light depth sensing, geometry-based monocular estimation, and language-aligned semantic prediction.

\paragraph{Evaluation Metrics.}
Since the target of SuperFace is to produce facial expressions that are more perceptually faithful to humans, we adopt human preference as the primary evaluation criterion.
For each test sample, we render two facial expressions driven by the coefficient predictions of the evaluated model and a baseline, respectively, and ask annotators to judge which rendered image better matches the reference image.

% We organize the metrics into two categories:
% \begin{itemize}
% \item \textbf{Human preference evaluation.} We report \textbf{Win Rate} and \textbf{Vote Ratio}. 
% %Win Rate measures the proportion of valid samples where the evaluated method is preferred. Vote Ratio denotes the ratio between the number of preferred votes for the evaluated method and that for the baseline. Self-Consistency measures the agreement rate under AB/BA order reversal.
% \item \textbf{Preference discriminator evaluation.} We report \textbf{Self-Consistency}, \textbf{2-Class Accuracy} (A/B), \textbf{3-Class Accuracy} (A/B/Similar), and \textbf{Macro-F1}.
% \end{itemize}
We organize the metrics into two categories: a) Human preference evaluation (Win Rate and Vote Ratio) and b) Preference discriminator evaluation (Self-Consistency, 2-Class Accuracy (A/B), 3-Class Accuracy (A/B/Similar), and Macro-F1). 
Detailed metric definitions and implementations are provided in the Appendix.

\paragraph{Implementation Details.}
We adopt Qwen3-VL-4B-Instruct\cite{qwen3vl} as the backbone model and use LoRA\cite{hu2021lora} for all training stages.
In the cold-start stage, the model is trained on the SFT split to establish a basic image-to-ARKit correspondence.
After initialization, we roll out candidate ARKit coefficients on the seven-subject split with sampling temperature $T=1.0$, and compare them with the corresponding Live Link Face coefficients to construct preference data.
We then train a preference discriminator on the resulting preference pairs and further optimize the policy model with DPO.

For all stages, we use bfloat16 training, a learning rate of $1\times10^{-4}$, a per-device batch size of 4, and DeepSpeed ZeRO-2.
LoRA is applied with rank 8 and $\alpha=32$, while the vision encoder and multimodal aligner are frozen.
% DPO training is terminated once the human preference win rate against Live Link Face exceeds a predefined threshold $\tau=0.8$ on the validation protocol.

% \paragraph{Implementation Details.}
% We adopt Qwen3-VL-4B-Instruct as the backbone model and use LoRA for all training stages.
% In the cold-start stage, the model is trained on the SFT split to establish a basic image-to-ARKit correspondence.
% After initialization, we roll out candidate ARKit coefficients on the seven-subject split with sampling temperature $T=1.0$, and compare them with the corresponding Live Link Face coefficients to construct preference data.
% We then train a preference discriminator on the resulting preference pairs and further optimize the policy model with DPO.

% For all stages, we use bfloat16 training, a learning rate of $1\times10^{-4}$, a per-device batch size of 4, and DeepSpeed ZeRO-2.
% LoRA is applied with rank 8 and $\alpha=32$, while the vision encoder and multimodal aligner are frozen.
% DPO training is terminated once the human preference win rate against Live Link Face exceeds a predefined threshold $\tau=0.8$ on the validation protocol.

% \subsection{Comparison with Existing Methods}
\subsection{Comparison of Human Expression Estimation}
\label{subsec:main_comparison}

% \begin{table}[h]
% \centering
% \caption{Human preference comparison against Live Link Face. Our method is the only approach that surpasses the 3D structured-light-based Live Link Face baseline.}
% \label{tab:main_results}
% \begin{tabular}{lcccc}
% \toprule
% Comparison &  & \multicolumn{3}{c}{Total Votes} \\
%  & Win Rate$\uparrow$ & Vote A$\uparrow$ & Vote B$\downarrow$ & Vote Tie \\
% \midrule
% DeadFace v.s Livelinkface     & 17.35\% & 34.7\% & 65.2\% & 0.1\% \\
% SemanticFace v.s Livelinkface & 46.15\% & 41.9\% & 41.3\% & 16.8\% \\
% \textbf{Ours v.s Livelinkface}         & \textbf{61.07\%} & \textbf{47.9\%} & \textbf{39.8\%} & \textbf{12.3\%} \\
% \bottomrule
% \end{tabular}
% \end{table}

\begin{table}[h]
\centering
\caption{Human preference comparison against Live Link Face(LLF). Our method is the only approach that surpasses the 3D structured-light-based Live Link Face baseline.}
\label{tab:main_results}
\begin{tabular}{lcccc}
\toprule
Comparison & Win Rate$\uparrow$ & \multicolumn{3}{c}{Total Votes} \\
\cmidrule(lr){3-5}
 &    & Method$\uparrow$ 
    & LLF$\downarrow$ 
    & Similar \\
\midrule
DeadFace v.s.LLF      & 17.35\% & 34.70\% & 65.20\% & 0.10\% \\
SemanticFace v.s. LLF  & 46.15\% & 41.90\% & 41.30\% & 16.80\% \\
Ours v.s. LLF & \textbf{61.07\%} & \textbf{47.90\%} & \textbf{39.80\%} & 12.30\% \\
\bottomrule
\end{tabular}
\end{table}

\begin{figure*}[h]
    \centering
    \includegraphics[width=\textwidth]{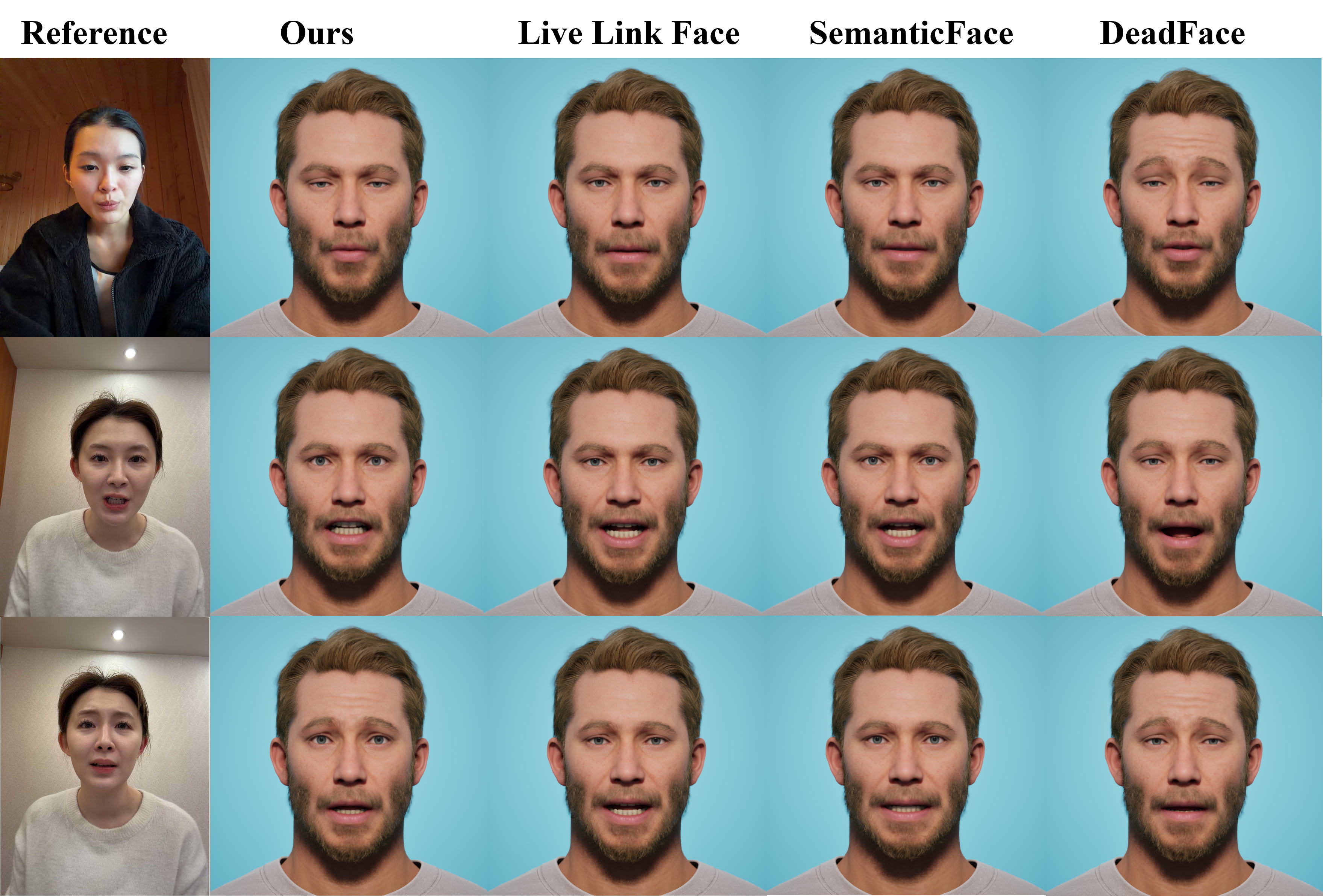}
    \caption{
    \textbf{Qualitative comparison of ARKit facial expression estimation.}
    SuperFace produces facial animations that are more perceptually consistent with the reference images, especially in local facial actions and overall expression semantics.
    }
    \label{fig:qualitative_comparison}
\end{figure*}

We compare SuperFace with DeadFace, SemanticFace, and Live Link Face on the held-out test set.
Live Link Face is a particularly strong baseline because it leverages 3D depth-camera sensing rather than pure 2D RGB input.
As shown in Tab.~\ref{tab:main_results}, DeadFace achieves only a 17.35\% win rate against Live Link Face, while SemanticFace improves the win rate to 46.15\%, approaching parity.

SuperFace achieves a win rate of 61.07\%, making it the only evaluated method that surpasses Live Link Face in human preference.
As shown in Fig.~\ref{fig:qualitative_comparison}, SuperFace better preserves local facial actions, such as eye/eyebrow motions and mouth shapes, and produces global expressions that are more consistent with the reference images.

This is notable because SuperFace uses only 2D RGB input, whereas Live Link Face benefits from depth sensing.
The improvement mainly comes from DPO-based preference optimization, which directly optimizes perceptual quality beyond geometric or semantic fidelity alone.
These results demonstrate that iterative preference feedback can effectively improve ARKit coefficient prediction and offers potential for further refinement with more preference data.

% \subsection{Effectiveness of Preference Discriminator}
\subsection{Analysis of Human Preference Discriminator}
\label{subsec:discriminator_eval}
Given a reference image and two candidates, determining which candidate better matches the reference is inherently subjective, especially when differences involve subtle facial actions. Human self-consistency on this task is only 61.3\%, so we adopt an AB/BA cross-validation and a multi-voter mechanism to obtain more reliable labels for training and evaluation. Repeated human annotation is costly and time-consuming, highlighting both the intrinsic difficulty of the task and the value of a trained discriminator as a cost-effective alternative.

We compare three discriminator designs: \textbf{Embed-based}~\cite{vemulapalli2019compact}, which estimates preference from image embedding similarity; \textbf{397B-MLLM} (Qwen3.5-397B-A17B\cite{qwen3vl}), which uses a large off-the-shelf multimodal model for direct preference judgment without task-specific training; and \textbf{Trained 4B-MLLM} (Qwen3-VL-4B-Instruct \cite{qwen3vl}), which is fine-tuned from 1000 human preference labels. 
The performance of each discriminator on the hard task is summarized in Tab.~\ref{tab:discriminator_results}. 
The embed-based discriminator performs poorly across all metrics, indicating that raw visual features are insufficient for capturing fine-grained facial actions.
The off-the-shelf 397B-MLLM, despite its strong general reasoning capability, tends to give "similar" answers rather than making clear preference decisions; as a result, although its raw accuracy appears acceptable, its Macro-F1 is low because most correct predictions come from the "similar" option rather than genuine preference discrimination.
In contrast, the trained 4B-MLLM achieves the best balance between prediction accuracy and robustness, nearly approaching human-level performance in 2-class accuracy and Macro-F1. 
Additional comparisons on discriminator performance are reported in the Appendix.

% \begin{table}[h]
% \centering
% \caption{Performance of preference discriminators on the hard task. The trained 4B-MLLM achieves the best balance between accuracy and robustness, approaching human-level performance.}
% \label{tab:discriminator_results}
% \resizebox{\textwidth}{!}{
% \begin{tabular}{lcccc}
% \toprule
% Method & Self-consistency $\uparrow$ & 2-class acc. $\uparrow$ & 3-class acc. $\uparrow$ & MacroF1 $\uparrow$ \\
% \midrule
% Embed-based          & ---  & 54.4\% & ---   & 0.18 \\
% 397B-MLLM            & 51.3\% & 60.0\% & 58.4\% & 0.34 \\
% Trained 4B-MLLM      & 52.0\% & 67.4\% & 59.3\% & 0.47 \\
% Human-to-human       & 61.3\% & 78.0\% & 63.3\% & 0.46 \\
% \bottomrule
% \end{tabular}
% }
% \end{table}

\begin{table}[h]
\centering
\caption{Performance of preference discriminators on the hard task. Human-to-human agreement is reported as a reference and is not included in the model comparison. The trained 4B-MLLM achieves the best overall performance among automatic discriminators.}
\label{tab:discriminator_results}
\resizebox{\textwidth}{!}{
\begin{tabular}{lcccc}
\toprule
Method & Self-consistency $\uparrow$ & 2-class acc. $\uparrow$ & 3-class acc. $\uparrow$ & MacroF1 $\uparrow$ \\
\midrule
Human-to-human       & 61.3\% & 78.0\% & 63.3\% & 0.46 \\
\midrule
Embed-based          & ---    & 54.4\% & ---    & 0.18 \\
397B-MLLM            & 51.3\% & 60.0\% & 58.4\% & 0.34 \\
Trained 4B-MLLM      & \textbf{52.0\%} & \textbf{67.4\%} & \textbf{59.3\%} & \textbf{0.47} \\
\bottomrule
\end{tabular}
}
\end{table}

\subsection{Analysis of Iterative Preference Optimization}
\label{subsec:dpo_iter}

Starting from the cold-start SFT model, we apply iterative DPO to progressively refine the policy model.
At each iteration, the updated model generates candidate ARKit coefficients, which are compared with the corresponding Live Link Face coefficients to construct preference pairs.
A trained preference discriminator then labels these pairs, and the policy model is further optimized using the resulting chosen-rejected data.
Tab.~\ref{tab:dpo_iteration} reports both human preference and discriminator results across iterations.
The SFT model, trained to imitate Live Link Face coefficients, achieves a low win rate of 16.16\% against the baseline.
After the first DPO round, the win rate increases substantially to 46.43\% in the user study. 
By round 2, the human preference win rate peaks at 61.07\%, demonstrating that iterative visual feedback effectively shifts the policy model toward more perceptually faithful facial actions.

However, the improvement does not continue monotonically.
In round 3, the human preference win rate drops to 41.26\%, while the discriminator remains high at 87.87\%.
This divergence indicates that the policy begins to overfit to the discriminator's judgment on the training data rather than generalizing to genuine human preference.
Consequently, we adopt the round-2 model as the final policy model.
A detailed analysis of the overfitting behavior is provided in the Appendix.

\begin{table}[h]
\centering
\caption{The win rate of iterative DPO optimization. The human preference win rate peaks at round 2 and declines in round 3 due to overfitting.}
\label{tab:dpo_iteration}
\begin{tabular}{lccc}
\toprule
        & User Study$\uparrow$ & Discriminator$\uparrow$ \\
\midrule
Round 0 (SFT)     & 16.16\%  & 32.42\% \\
Round 1 & 46.43\%  & 74.07\% \\
Round 2 & \textbf{61.07\%}  & \textbf{95.38\%} \\
Round 3 & 41.26\%  & 87.87\% \\
\bottomrule
\end{tabular}
\end{table}

\section{Conclusion}
\label{sec:conclusion}

In this paper, we proposed \textbf{SuperFace}, a preference-driven framework for ARKit facial expression estimation. SuperFace uses Live Link Face coefficients as initialization rather than fixed ground truth, and further improves coefficient prediction through region-aware preference construction and DPO optimization. By aligning ARKit prediction with perceptual judgments, SuperFace produces more faithful facial expressions and demonstrates the potential of preference-driven optimization for digital human animation.

\paragraph{Limitations.}Although SuperFace improves ARKit coefficient prediction through preference-driven optimization, it still has several limitations. First, the preference discriminator relies on initial human annotations, which introduce additional labeling cost and may contain subjective bias. Second, our region-aware construction only divides the face into upper and lower regions, and may not fully capture fine-grained local motions or asymmetric expressions. Third, preference comparison depends on rendered images, so the renderer and avatar quality may influence the collected preferences. Future work may explore more fine-grained region decomposition, stronger automatic preference models, and broader generalization across identities, renderers, and facial animation systems. Beyond technical limitations, SuperFace may benefit digital human animation, virtual production, and AR/VR by improving facial expression fidelity, but more realistic facial animation may also be misused for deceptive synthetic media. We therefore encourage consent-based use and clear disclosure of synthetic content.
\newpage

\bibliographystyle{abbrvnat}  %plainnat,abbrvnat,unsrtnat
\small
\bibliography{ref}
\clearpage

\appendix

\begin{center}
{\Large \textbf{Appendix}}

  % Supplementary Material 
\end{center}

\section{Evaluation Metrics Details}

We use the following metrics to evaluate model performance:

\noindent\textbf{Total votes.} The total number of valid votes collected in the user study, reported in A:B:Similar format, represents votes for method A, method B, and similar, respectively. 

\noindent\textbf{Win Rate.}
We compute the win rate through a pairwise user study comparing the target method against Live Link Face.
For each comparison, two annotators are asked to choose among ``A'', ``B'', and ``Similar''.
To improve reliability, each pair is evaluated twice with randomized display order, where the A/B positions are swapped for cross-validation.
We retain only high-confidence non-similar comparisons, i.e., cases where both annotators consistently prefer the same method under both display orders, resulting in four consistent votes.
The win rate is then computed as the proportion of valid comparisons in which the target method is preferred over Live Link Face:
\begin{equation}
\mathrm{WinRate}
=
\frac{N_{\mathrm{win}}}
{N_{\mathrm{win}} + N_{\mathrm{lose}}},
\end{equation}
where $N_{\mathrm{win}}$ and $N_{\mathrm{lose}}$ denote the numbers of high-confidence four-vote comparisons won and lost by the target method, respectively.

where $N_{\mathrm{win}}$ and $N_{\mathrm{lose}}$ denote the numbers of valid samples where the evaluated method wins or loses against the baseline. 

\noindent\textbf{Self-consistency.} To eliminate position bias, we perform two inference passes per sample: first with (A, B) order, then with swapped (B, A) order. A sample is counted as self-consistent if both passes select the same image. Self-consistency is defined as the proportion of self-consistent samples, reflecting prediction stability.

\noindent\textbf{2-class accuracy.} Binary classification accuracy on A vs. B options only. Predictions matching ground truth are correct; otherwise incorrect. Similar and invalid predictions are excluded from this metric.

\noindent\textbf{3-class accuracy.} Ternary classification accuracy on A, B, and similar options. Predictions must exactly match ground truth to be correct. This metric directly reflects the model's overall discrimination capability in the three-class setting.

\noindent\textbf{MacroF1 (A/B/Similar).} Macro-averaged F1 score for the three-class task, computed as the arithmetic mean of per-class F1 scores. Reported as (F1$_A$/F1$_B$/F1$_T$ - MacroF1), where the first three values are per-class F1 and the last is the macro average. This metric better reflects comprehensive discrimination under class imbalance than accuracy.

\section{Comparison of Preference Discriminator}

\begin{table}[h]
\centering
\caption{Comparison of preference discriminators on Easy and Hard tasks.}
\label{tab:discriminator_results}
\resizebox{\textwidth}{!}{
\begin{tabular}{lcccccc}
\toprule
\textbf{Method} & \textbf{Self-consistency} & \textbf{2-class acc.} & \textbf{Self-consistency} & \textbf{2-class acc.} & \textbf{3-class acc.} & \textbf{MacroF1} \\
 & (Easy) & (Easy) & (Hard) & (Hard) & (Hard) & (Hard) \\
\midrule
Embedded Judger & - & 64.0\% & - & 54.4\% & - & (0.30/0.24/0.00 -- 0.18) \\
397B-MLLM Judger & 89.4\% & 94.0\% & 51.3\% & 60.0\% & 58.4\% & (0.17/0.13/0.74 - 0.34) \\
Trained 4B-MLLM Judger & $\sim$100.0\% & $\sim$100.0\% & 52.0\% & 67.4\% & 59.3\% & (0.41/0.46/0.54 -- 0.47) \\
Human-to-human & $\sim$100.0\% & $\sim$100.0\% & 61.3\% & 78.0\% & 63.3\% & (0.62/0.61/0.13 -- 0.46) \\
\bottomrule
\end{tabular}
}
\end{table}

We evaluate the preference discriminator on two tasks with varying difficulty. The Easy Task is a binary classification (A/B) on pairs with large expression differences (top 10\% by $MSE =\|\mathcal{S}_A - \mathcal{S}_B\|^2$), while the Hard Task is a ternary classification (A/B/Similar) on unfiltered data containing near-identical expressions. The result is shown in Tab. ~\ref{tab:discriminator_results}.

The embedded judger fails to provide fine-grained discrimination---it can classify coarse expressions but cannot distinguish subtle differences. The 397B-MLLM achieves 94.0\% 2-class accuracy on the Easy Task but struggles on the Hard Task, exhibiting a strong bias toward the ``similar'' label for ambiguous cases rather than discriminating subtle facial differences. This reflects a misalignment between its pretrained ``similar'' threshold and our task requirements. Our fine-tuned 4B-MLLM judger achieves 59.3\% 3-class accuracy with MacroF1 of 0.47, approaching human-level performance (around 60\%--70\%).

\section{Analysis on Region-Aware Construction}

\begin{table}[h]
\centering
\caption{Ablation study on region-aware (R.A.) construction strategy.}
\label{tab:region-aware}
\resizebox{\textwidth}{!}{
\begin{tabular}{llcccc}
\toprule
\textbf{Training Data} & \textbf{Inference Task} & \textbf{Self-consistency} & \textbf{2-class acc.} & \textbf{3-class acc.} & \textbf{MacroF1} \\
\midrule
W/O R.A. & W/O R.A. & 68.0\% & 67.1\% & 41.0\% & (0.41/0.31/0.10 -- 0.27) \\
W/ R.A. & W/O R.A. & 52.0\% & 67.4\% & 59.3\% & (0.41/0.46/0.54 -- 0.47) \\
W/ R.A. & W/ R.A. & 83.3\% & 93.5\% & 71.1\% & (0.81/0.69/0.62 -- 0.70) \\
\bottomrule
\end{tabular}
}
\end{table}

We introduce a region-aware preference construction strategy that decomposes full-face comparison into upper-face and lower-face judgments, enabling more reliable pairwise preference annotation.

Comparing the first and third rows in Tab.~\ref{tab:region-aware}, region-aware training and inference significantly outperform the full-face baseline across all metrics. The full-face approach suffers from reduced annotation quality, as observing both regions simultaneously confuses human annotators, and inference is inherently more difficult without region isolation. Comparing the second and third rows, region-aware training generalizes well to full-face inference. While self-consistency drops due to task mismatch, the model maintains strong accuracy on answered samples, demonstrating that bidirectional verification ensures reliable predictions even when training and inference tasks differ.

\section{Overfitting and Stopping in Iterative Preference Optimization}

\begin{table}[h]
\centering
\caption{Iterative DPO optimization results. Vote format is Win:Lose:Tie. The human preference win rate peaks at round 2 and declines in round 3 due to overfitting.}
\label{tab:dpo_iteration2}
\begin{tabular}{lcccc}
\toprule
        & \multicolumn{2}{c}{User Study} & \multicolumn{2}{c}{Discriminator-4B} \\
        \cmidrule(lr){2-3}\cmidrule(lr){4-5}
        & Win Rate$\uparrow$ & Total Votes & Win Rate$\uparrow$ & Total Votes \\
\midrule
SFT     & 16.16\%  & 308:676:16  & 32.42\% & 134:201:265 \\
Round 1 & 46.43\%  & 464:500:36  & 74.07\% & 218:114:268 \\
Round 2 & 61.07\%  & 479:398:123 & 95.38\% & 365:44:191  \\
Round 3 & 41.26\%  & 381:456:163 & 87.87\% & 366:95:139  \\
\bottomrule
\end{tabular}
\end{table}

In Tab.~\ref{tab:dpo_iteration2}, the overfitting phenomenon observed at round 3 stems from an inherent limitation of using a learned discriminator for preference labeling. Specifically, although the preference discriminator is trained to approximate human judgment, it exhibits systematic deviations from true human preferences. When evaluating generated image pairs, these deviations manifest as consistent biases toward certain facial expression patterns. As DPO training proceeds iteratively, the model increasingly exploits these biases, learning to generate outputs that the discriminator favors but that may not align with actual human preferences. This is evidenced by the continuous decline in "similar" votes from the model judge, as the discriminator becomes more confident in distinguishing the increasingly biased outputs.

Also, despite the drop in win rate at round 3, the proportion of "similar" votes in the user study continued to increase. We attribute this trend to the enhanced quality of our half-face rendering, which in some cases surpasses Live Link Face in subtlety and realism. However, some biases favored by the discriminator are not recognized by humans. This ambiguity, arising specifically from the half-face condition, introduces additional difficulty during human evaluation, making it more challenging for annotators to confidently distinguish between methods, thereby increasing the proportion of "similar" judgments. This observation further supports our decision to halt training at round 2, where the model achieved an optimal balance between preference optimization and generalization.

\section{Analysis on Training Strategy}

\begin{table}[h]
\centering
\caption{Analysis on training strategy.}
\label{tab:trainig_strategy}
\begin{tabular}{lcc}
\toprule
\textbf{Method} & \textbf{Total votes} & \textbf{Win rate} \\
\midrule
User Study + DPO & 441:564:195 & 34.52\% \\
Discriminator + DPO & 538:515:147 & 53.60\% \\
\bottomrule
\end{tabular}
\end{table}

Due to the scarcity and high cost of human annotations, we train a preference discriminator on 1,000 human-labeled samples to generate DPO training data. We compare this against directly using the same human data for DPO, shown in Tab. ~\ref{tab:trainig_strategy}. 

Direct DPO training on limited human data leads to severe overfitting, resulting in a win rate of only 34.52\% in user study. In contrast, using the same human data to train a preference discriminator is significantly more efficient, achieving a 53.60\% win rate. This demonstrates that learning a preference oracle provides better data efficiency than direct policy optimization under human annotation constraints.

\normalsize
% \clearpage
% \newpage
% \input{checklist.tex}

\end{document}